\documentclass[a4paper,fleqn]{cas-dc}
\usepackage{amsmath,amsfonts,amsthm}
\usepackage{algorithmic}
\usepackage{algorithm}
\usepackage{array}
\usepackage{upgreek}
\usepackage{textcomp}
\usepackage{stfloats}
\usepackage{url}
\usepackage{verbatim}
\usepackage{graphicx}
\usepackage{multirow}
\usepackage{caption}
\usepackage{hyperref}
\usepackage[numbers]{natbib}
\def\tsc#1{\csdef{#1}{\textsc{\lowercase{#1}}\xspace}}
\tsc{WGM}
\tsc{QE}
\makeatletter
\renewcommand{\fnum@figure}{Fig. \thefigure.\@gobble}
\makeatother

\begin{document}
\let\WriteBookmarks\relax
\def\floatpagepagefraction{1}
\def\textpagefraction{.001}
\let\printorcid\relax
\shorttitle{An Intelligent Social Learning-based Optimization Strategy for Black-box Robotic Control with Reinforcement Learning}    

\shortauthors{Xubo Yang et~al.}  

\title [mode = title]{An Intelligent Social Learning-based Optimization Strategy for Black-box Robotic Control with Reinforcement Learning}

\author[1]{Xubo Yang}

\author[1]{Jian Gao}
\cormark[1]

\author[2]{Ting Wang}

\author[1]{Yaozhen He}

\affiliation[1]{organization={School of Marine Science and Technology, Northwestern Polytechnical University},
            city={Xi'an},     
            postcode={710072}, 
            state={Shaanxi},
            country={China}}

\affiliation[2]{organization={College of Electrical Engineering And Control Science, Nanjing Tech University},
            city={Nanjing},
            postcode={211816}, 
            state={Jiangsu},
            country={China}}

\begin{abstract}
 Implementing intelligent control of robots is a difficult task, especially when dealing with complex black-box systems, because of the lack of visibility and understanding of how these robots work internally. This paper proposes an Intelligent Social Learning (ISL) algorithm to enable intelligent control of black-box robotic systems. Inspired by mutual learning among individuals in human social groups, ISL includes learning, imitation, and self-study styles. Individuals in the learning style use the Levy flight search strategy to learn from the best performer and form the closest relationships. In the imitation style, individuals mimic the best performer with a second-level rapport by employing a random perturbation strategy. In the self-study style, individuals learn independently using a normal distribution sampling method while maintaining a distant relationship with the best performer. Individuals in the population are regarded as autonomous intelligent agents in each style. Neural networks perform strategic actions in three styles to interact with the environment and the robot and iteratively optimize the network policy. Overall, ISL builds on the principles of intelligent optimization, incorporating ideas from reinforcement learning, and possesses strong search capabilities, fast computation speed, fewer hyperparameters, and insensitivity to sparse rewards. The proposed ISL algorithm is compared with four state-of-the-art methods on six continuous control benchmark cases in MuJoCo to verify its effectiveness and advantages. Furthermore, ISL is adopted in the simulation and experimental grasping tasks of the UR3 robot for validations, and satisfactory solutions are yielded.
\end{abstract}



\begin{keywords}
  Intelligent social learning \sep Robot control \sep Black-box optimization \sep Reinforcement learning \sep Intelligent optimization 
\end{keywords}

\maketitle

\section{Introduction}
A black-box model is characterized by opaque internal decision-making processes and operational mechanisms that are challenging for outsiders to comprehend. It can be likened to a sealed box, where only inputs and outputs are observable and understandable \cite{dong2020surrogate}. In the pursuit of intelligent control of complex robotic systems, these systems are frequently treated as black-box models. In this approach, the focus is solely on the input states and output actions of the robot, with no consideration given to the internal motion mechanisms \cite{qin2022sablas}. Developing agents that can accomplish challenging tasks in complex, uncertain robotic systems is a key goal for black-box intelligent control.

In recent years, Reinforcement Learning (RL) has emerged as the most widely used approach to intelligent control of complex robotic systems. The basic idea of RL is that the agent continually adapts its strategies while interacting with the environment to maximize rewards \cite{chen2023deep,zhu2021survey,arulkumaran2017deep}. Numerous fields, including Atari games \cite{mnih2016asynchronous}, GO \cite{silver2016mastering}, and robotic control tasks \cite{ramirez2023reinforcement,elguea2023review,liu2022robot}, have demonstrated the effectiveness of various RL-related algorithms. The synthesis of RL with Deep Learning is commonly referred to as Deep Reinforcement Learning (DRL). DRL exhibits formidable perception, representation, and decision-making capabilities, enabling it to address scenarios characterized by highly complex state and action spaces. Deep Deterministic Policy Gradient (DDPG) is an algorithm that operates based on deterministic policies. It has been successfully applied to obstacle avoidance problems in robotics \cite{wang2023research}, as well as to manipulative tasks involving robotic arm grasping \cite{shao2023control} and \cite{dong2023enhanced}. In contrast, Soft Actor-Critic (SAC) employs stochastic policies and has been applied to mobile robot path planning \cite{yu2022self} and \cite{wang2022autonomous}, as well as robotic arm motion control \cite{acuto2022variational}. Proximal Policy Optimization (PPO) is an on-policy algorithm that has been utilized for gait control in humanoid robots \cite{kuo2023intelligent}.

Intelligent Optimization (IO) has emerged as a prominent research direction in the fields of intelligent science, information science, and artificial intelligence for addressing black-box optimization problems \cite{li2021survey,wang2001intelligent}. IO, a random search algorithm, draws inspiration from biological and natural phenomena. It simulates behaviors observed in certain social species, such as foraging and breeding, and abstracts these actions into quantifiable key indicators. Mathematical models are then developed to solve a wide range of problems. With its advantages of parallelism, versatility, and robustness, IO provides a novel approach for finding optimal solutions, particularly in situations where models are either unavailable or challenging to create. However, in the domain of robotics, current research efforts predominantly focus on utilizing IO algorithms for path planning \cite{wu2023modified,liu2023improved,dai2023novel,qu2020novel} or combining IO algorithms with traditional control methods for motion control \cite{singhal2022robust,hasan2022disturbance,ghith2022design}.

In addition, another prevalent area of research focuses on applying IO algorithms to optimize neural networks for achieving intelligent control in robotics \cite{zhou2021survey}. A noteworthy approach in this regard is NeuroEvolution, which utilizes evolutionary algorithms to generate and optimize neural networks. NeuroEvolution offers significant advantages in handling highly complex, nonlinear, and high-dimensional problems, eliminating the need for manual parameter configuration of the network \cite{miller1989designing,stanley2002evolving,salimans2017evolution}.

Although DRL and IO can deal with complex black-box problems and have been successfully applied to many challenging domains, they still have shortcomings. The agents in DRL are extremely sensitive to hyperparameters, implementation details, and uncertainties in environmental dynamics. At the same time, the gradient update needs to be repeated in the training process, which leads to time-consuming calculation, and there are also problems of temporal credit assignment with long time horizons and sparse rewards. In contrast, IO does not care about rewards distribution, does not need backpropagating gradients, and has a better exploration mechanism. However, in real-time control systems where decisions must be made quickly, IO must spend time searching for the best solution for each state in the control field. Additionally, IO is perceived as less effective at solving complex high-dimensional problems and easily falling into local optimum. 

Addressing the advantages and limitations of RL and IO, we propose Intelligent Social Learning (ISL) as an innovative and sophisticated intelligent optimization algorithm. ISL is based on IO and draws inspiration from the policy learning approach in RL, enabling it to effectively handle complex black-box optimization control problems. ISL imitates mutual learning among individuals observed in human social groups, and includes three styles, namely learning style, imitation style, and self-study style. In the learning style, individuals acquire knowledge from the best performer, establish the closest relationship with the best, and have the largest population. In the imitation style, individuals mimic the best performer, have a second-level rapport with the best, and constitute the second-largest population. In the self-study style, individuals learn independently, maintain a distant relationship with the foremost performer, and comprise the smallest population. The main contribution of this paper can be summarized as below:
\begin{itemize}
	\item{A novel algorithm named Intelligent Social Learning (ISL) is proposed for controlling robots, and it can significantly reduce the computational cost of the optimization process.}
	\item{Comparing ISL with DDPG, SAC, PPO, and EA on six continuous control benchmark cases in MuJoCo, the results show that ISL has its characteristics and advantages and has certain research value and promotion significance.}
	\item{Design the grasping task of the UR3 robot, and apply ISL to the simulation and experimental testing of the robot, achieving satisfactory results.}
\end{itemize}

The rest of the present study is organized as follows. Section 2 introduces the theoretical background of the Markov Decision Process and Lévy flight. The proposed ISL algorithm is explained in detail in Section 3. In Section 4, ISL is tested through experiments and used in engineering applications. Lastly, the conclusions are drawn in Section 5.

\section{Theoretical Background}
\subsection{Markov Decision Process}
The Markov property refers to a stochastic process where the conditional probability distribution of future states depends only on the present state and not on any past states, given the present state. When a system's environment has the Markov property, simulating an agent's interaction with the environment to achieve a stochastic policy and return is called a Markov Decision Process (MDP)\cite{kurniawati2022partially}. As a result, optimization problems like driving control that can be resolved through dynamic programming (i.e., DRL) can be handled effectively with MDP. Applying MDP in the domain of IO to search for optimal control strategies is a primary modeling approach of this paper.

The MDP can be defined with the tuple $\left( {S,A,P,R,\gamma } \right)$, which consists of a state space $S$, an action space $A$, a transition probability model $P$, a reward function $R$, and a discounter factor $\gamma  \in \left( {0,1} \right]$. In MDP, the agent accomplishes an action $a$ and then receives a scalar reward $r$ return from the environment according to the given reward function $R$. More specifically, Eq. (1) shows that the value of being in a state $s$ and adhering to the policy $\pi \left (s\right )$ until the episode's conclusion is defined by the value function ${V_{\pi}}$:

\begin{equation}
	{V_\pi }(s) = {\rm E}[{r_{t + 1}} + \gamma {r_{t + 2}} + {\gamma ^2}{r_{t + 3}} +  \cdots  + {\gamma ^{T - t - 1}}{r_T}\left| {{s_t} = s} \right.]
\end{equation}
where $s_t$ represents the state at time $t$.

According to the Bellman equation, its recursive expression is:
\begin{equation}
	{V_\pi }(s) = \sum\limits_{a \in A} {\pi (a\left| s \right.)} \left( {R(s,a) + \gamma \sum\limits_{s' \in S} {p(s'\left| {s,a} \right.)} {V_\pi }(s')} \right)
\end{equation}
where $\pi (a\left| s \right.)$ represents the probability of acting $a$ in state $s$, and $p(s'\left| {s,a} \right.)$ represents the probability of transitioning to state $s'$ after executing action $a$ in state $s$.

\subsection{Lévy Flight}
Lévy flight is a stochastic model of random walks commonly used to describe the movement of particles or animals in space \cite{kaidi2022dynamic,li2022survey}. The distinctive characteristic of Lévy flight is its ability to exhibit a wide range of step lengths, allowing it to cover large distances in relatively short periods. As a result, an increasing number of studies have applied Lévy flight to optimization problems \cite{yang2010engineering,wang2022novel,joshi2023levy}. In optimization algorithms, incorporating the movement pattern of Lévy flight can aid in the search for optimal solutions, particularly when faced with large search spaces and multiple local optima. Lévy flight effectively enhances sample diversity and search range, facilitating escape from local optima and enabling the discovery of superior solutions. Consequently, Lévy flight has emerged as a powerful optimization tool widely employed across various domains. The mathematical model of Lévy flight can be expressed simply by Eqs. (3) and (4).

\begin{equation}
	L\acute{e}vy \sim u = {t^{ - \lambda }},{\rm{  }}1 < \lambda \le 3
\end{equation}

\begin{equation}
	{L\acute{e}vy(s,\gamma ,\mu )} = \begin{cases}
		\sqrt{\frac{\gamma }{2\pi }} \exp \left [ - \frac{\gamma }{2\left ( s- \mu  \right ) }  \right ]\frac{1}{\left ( s- \mu  \right )^{\frac{3}{2} }}\vspace{1ex} ,{0< \mu < s< \infty} \\ 
		{0},{s\le 0.} 
	\end{cases}
\end{equation}
where $s$ represents the step size, $\mu$ is the displacement parameter, and $\gamma$ is the scale parameter (($\gamma > 0$, control distribution scale).

The Mantegna algorithm, described by Eqs. (5) and (6), is typically used to approximate the Lévy distribution, which is difficult to achieve due to its complexity.

\begin{equation}
	\lambda  = \frac{u}{{|v{|^{{\textstyle{1 \over \beta }}}}}}
\end{equation}
\begin{equation}
	{\sigma _u} = {\left\{ {\frac{{\Gamma (1 + \beta )\sin ({{\pi \beta } \mathord{\left/
							{\vphantom {{\pi \beta } 2}} \right.	\kern-\nulldelimiterspace} 2})}}{{{{\Gamma \left[ {{{(1 + \beta )} \mathord{\left/{\vphantom {{(1 + \beta )} 2}} \right.
										\kern-\nulldelimiterspace} 2}} \right]} \mathord{\left/	{\vphantom {{\Gamma \left[ {{{(1 + \beta )} \mathord{\left/{\vphantom {{(1 + \beta )} 2}} \right.
													\kern-\nulldelimiterspace} 2}} \right]} {\beta  \times {2^{{{(\beta  - 1)} \mathord{\left/
														{\vphantom {{(\beta  - 1)} 2}} \right.
														\kern-\nulldelimiterspace} 2}}}}}} \right.
							\kern-\nulldelimiterspace} {\beta  \times {2^{{{(\beta  - 1)} \mathord{\left/{\vphantom {{(\beta  - 1)} 2}} \right.	\kern-\nulldelimiterspace} 2}}}}}}}} \right\}^{{1 \mathord{\left/	{\vphantom {1 \beta }} \right.\kern-\nulldelimiterspace} \beta }}},{\rm{  }}{\sigma _v} = 1	
\end{equation}
where $u\sim N(0,\sigma _u^2)$, $v\sim N(0,\sigma _v^2)$, $u$ and $v$ follow a standard normal distribution, the power law index $\beta = 1.5$, and $\Gamma$ represents the conventional gamma function. 

The following equations also update the flight distances during the iteration.

\begin{equation}
	{x_i}(t + 1) = {x_i}(t) + \alpha  \oplus L\acute{e}vy(\lambda),{\rm{ }}i = 1,2,...,n	
\end{equation}
where ${x_i}(t)$ represents the current position, and ${x_i}(t + 1)$ represents the new position. $\alpha$ is the proportional factor of the step size, which should be related to the scales of the problem of interests. $\oplus$ is dot-multiplication, and $l\acute{e}vy(\lambda )$ presents the random search path.

\section{Intelligent Social Learning}
\subsection{Social Learning Theory}
Albert Bandura and Richard Walters, psychologists, proposed in the 1950s that individuals learn social knowledge, experience, and behavior by observing and imitating other individuals. During the learning process, mentoring one another and exchanging knowledge and perspectives help people learn faster and more effectively, according to the theory. In addition, people tend to utilize historical information better because learners can observe each other's actions and consequences and make decisions based on these observations.

The ISL algorithm draws on social learning theory and takes students and teachers in teaching as an illustration object. Its specific process includes three ways of knowledge acquisition: learning style, imitation style, and self-study style, as shown in Fig. 1. Similar to how most students in a class learn from the teacher with the most knowledge, the learning style is that individuals learn from the best members of the population. The teacher teaches the existing knowledge to students. Imitation style is when individuals imitate the learning skills of the best in the population, similar to how some students in a class imitate the teacher's problem-solving skills. Students can quickly get the correct results by using this skill and thinking. Self-study style is when individuals master learning methods and problem-solving skills independently, similar to how a few students in a class learn independently according to their ideas and methods instead of listening to the knowledge and methods taught by the teacher. 
\begin{figure}[h]
	\centering
	\includegraphics[width=3.0 in]{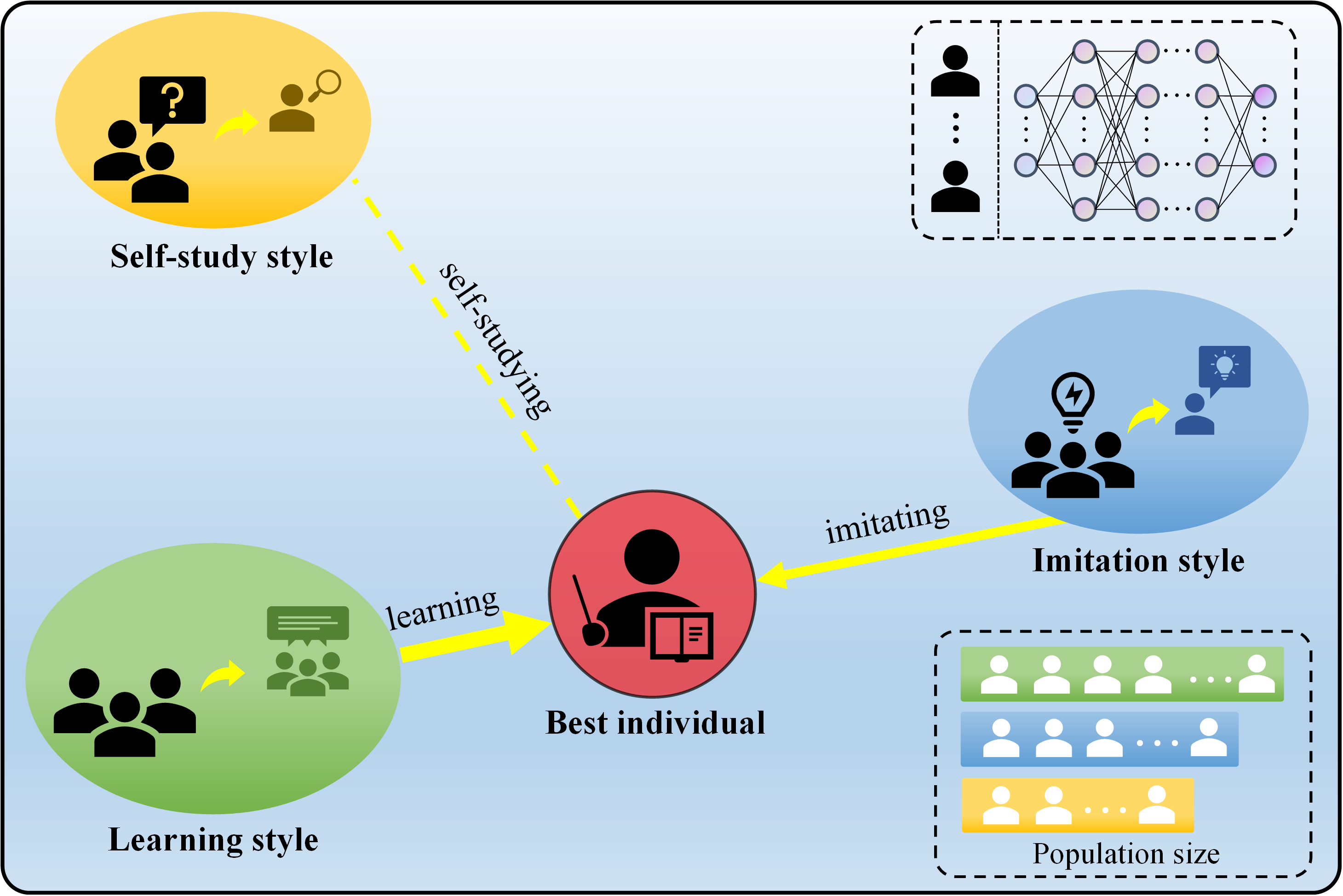}
	\caption{Intelligent social learning.}
	\label{fig_1}
\end{figure}

The number of individuals in the learning style accounts for the largest proportion of the total population, followed by the imitation style and then the self-study style, which aligns with the distribution proportion and learning rules of populations in human society. The individual in the population is represented as an independent agent composed of a deep neural network. For the convenience of subsequent description, the individuals in the population are collectively referred to as agents. The network structure of each agent is the same, but the parameters of weight and bias are different, indicating the difference in learning ability. The Markov decision process is used to assign the agents to distinct ISL styles per various proportions. Different styles produce distinct actions based on the state once the agent reaches a particular state in the environment. The action is executed, and the next state and the reward brought by the current action are output, the performance result of learning. All agents' model parameters are saved after the episode, and the agent with the highest historical performance is chosen as the best. The learning process of ISL is displayed in Fig. 2.

\begin{figure*}[h!]
	\centering
	\includegraphics[width=0.8\textwidth]{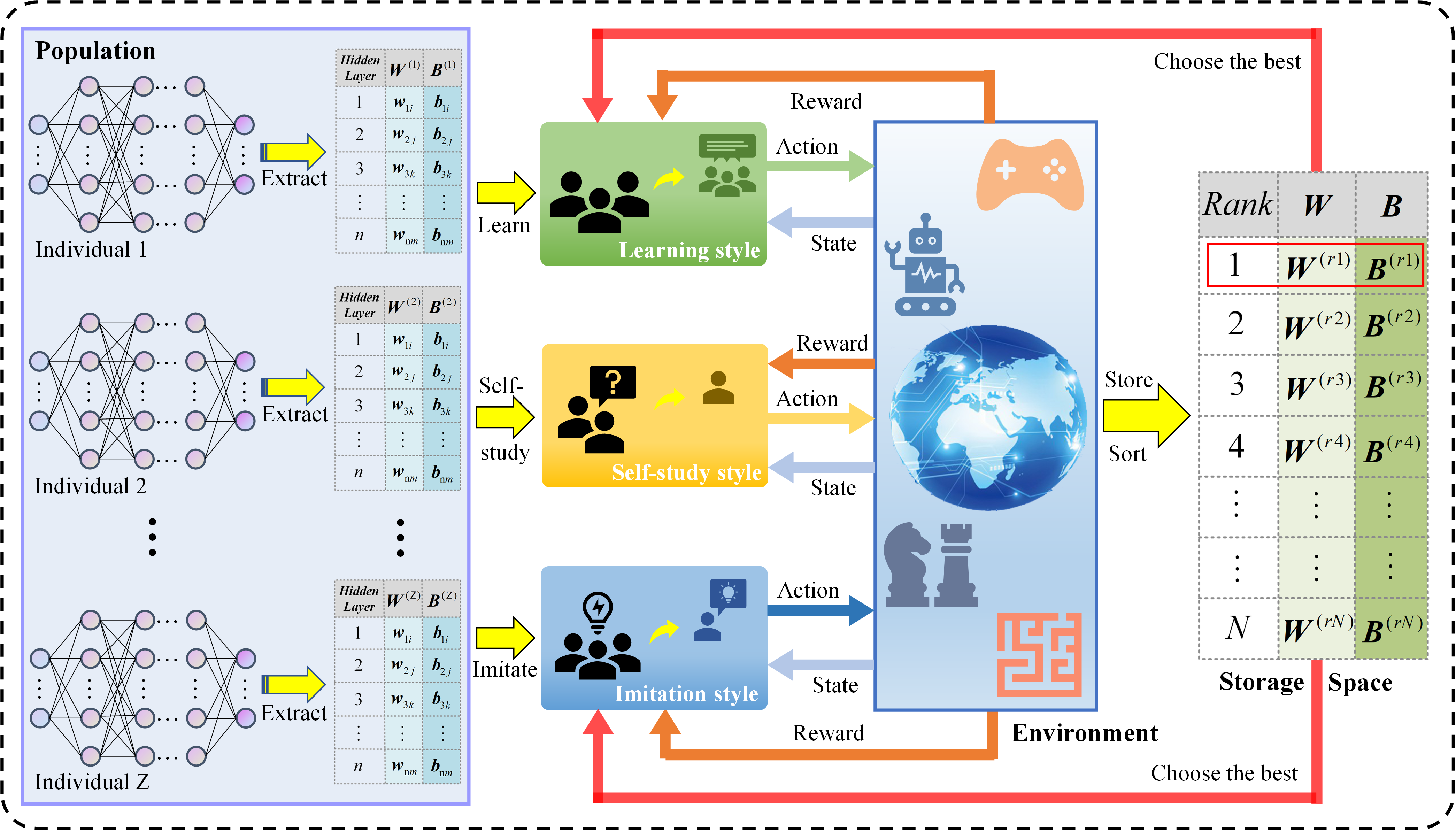}
	\caption{The learning process of ISL.}
	\label{fig_2}
\end{figure*}

\subsection{Three Strategic Styles}
As mentioned above, ISL includes three styles: learning, imitation, and self-study. Next, the parameters update process of the three styles will be further described in detail.

\subsubsection{Learning Style}
In the learning style of ISL, a Lévy flight search strategy is adopted to update each agent's position, with the flight distance's size representing the difference in learning ability between agents. Each agent in this style learns from the optimal solution and has the closest relationship with it. Eqs. (8) and (9), which describe the position update formula for Lévy flight, can be used to update the neural network parameters efficiently.

\begin{equation}
	{\theta _i}(t + 1) = {\theta _i}(t) + \alpha  \oplus L\acute{e}vy(\lambda ),{\rm{ }}i = 1,2,...,n	
\end{equation}
\begin{equation}
	\alpha  \oplus L\acute{e}vy(\lambda ) \sim \alpha \frac{u}{{|v{|^{ - \beta }}}}({\theta _i}(t) - {\theta _{{\rm{best}}}}(t))
\end{equation}
where ${\theta _i}(t)$ and ${\theta _i}(t + 1)$ represent the current agent's parameters (weights and bias) and the updated agent's parameters. ${\theta _{{\rm{best}}}}(t)$ represents the current optimal agent's parameters, and agent with the most knowledge, the historical optimal solution, is optimal.

Lévy flight in the position update, if the value of $\alpha$ is large, although the global search ability is enhanced, it is challenging to find high-precision solutions. Even though a small value of $\alpha$ increases the local search capacity, the algorithm will require multiple times as many iterations, reducing efficiency. This paper proposes a dynamic mechanism search in response to the above problems. The step size $\alpha$ changes from a fixed value to a dynamic step size with the number of iterations and $\alpha$ is expressed as follows:

\begin{equation}
	\alpha  = {\alpha _{\min }} + ({\alpha _{\max }} - {\alpha _{\min }})\cos {(\frac{\pi }{2} \cdot \frac{{step}}{{{max}\_step}})^2}	
\end{equation}
where $\alpha _{max} $ represents the maximum value of $\alpha$, and$\alpha _{min} $ represents the minimum value of $\alpha$. $step$ represents the iteration steps of the population, and ${max}\_step$ represents the population's maximum number of iteration steps.

The evolution of $\alpha$ during iteration is depicted in Fig. 3. In the initial iteration, it has a large value and slowly decays, increasing search capacity and preventing local optima. In the later iterations, $\alpha$ rapidly decreases to a smaller value, increasing the algorithm's local search capacity and facilitating rapid convergence.

\begin{figure}[h]
	\centering
	\includegraphics[width=3.3 in]{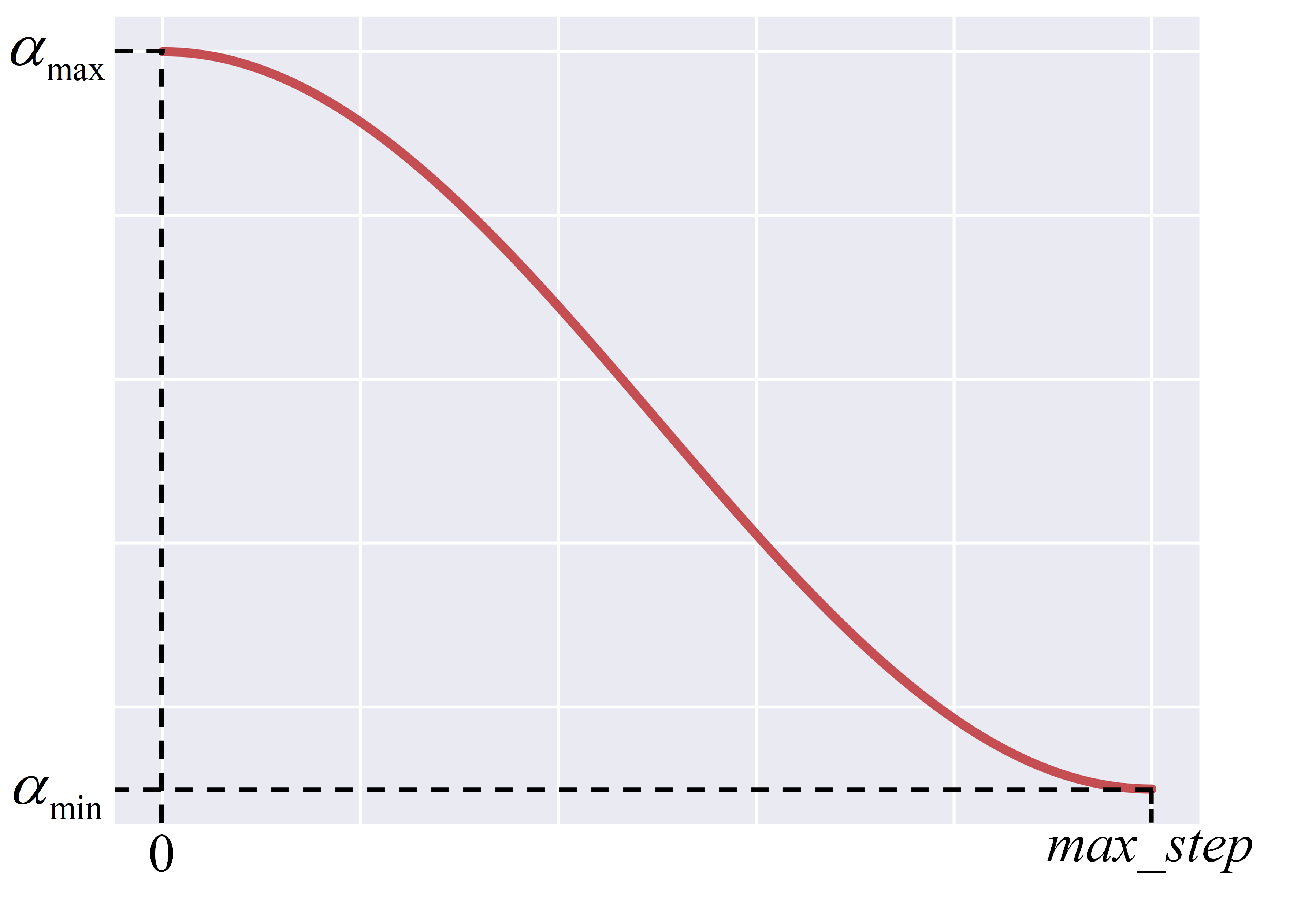}
	\caption{The change of $\alpha$  during iteration.}
	\label{fig_3}
\end{figure}

\subsubsection{Imitation Style}
A random disturbance strategy is used in the imitation style of ISL, with the magnitude of the disturbance representing the degree to which each agent's imitation result and the optimal solution are comparable. Each agent emulates the optimal solution in this style, and they have a second-close relationship with it. The imitation style randomly disturbs all parameters of the best agent in history with a 50$\%$ probability and some random parameters of the best agent with a 50$\%$ probability. The updated expression for disturbance is as follows:
\begin{equation}
	\left\{\begin{array}{l}\theta _{i} \left ( t+1 \right ) = \theta _{\text {best}}\cdot \left \{ 1+  \text{rand} \left ( a,b \right )  \right \} \vspace{0.5ex} \\,\theta = \theta _{i}\left ( t \right ) \& \text {Pr} \ge 0.5 \vspace{0.5ex}
		\\ \theta _{i} \left ( t+1 \right ) = \theta _{\text {best}}\left [ ind_{\text {1}} , ind_{\text {2}}\right ] \cdot \left \{ 1+  \text{rand} \left ( a,b \right )  \right \} \vspace{0.5ex}\\,\theta = \theta _{i}\left ( t \right ) \& \text {Pr} <  0.5. \vspace{0.5ex} 
	\end{array}\right.
\end{equation}
where $a$ and $b$ are the lower and upper limits of the disturbance amplitude, $ind_{1}$ and $ind_{2}$ are random numbers between the total number of parameters.

\subsubsection{Self-study Style}
In the self-study style of ISL, a normal distribution sampling strategy is adopted, with the sampling size representing the degree of self-study results for each agent. In this style, each agent learns independently and has the least connection to the best solution. The sampling strategy is independent of each layer of the network. This means that the mean and variance of each layer's distributions are the expected values and variance of the current layer's network parameters in the historical optimal solution, respectively. The updated expression for sampling is:

\begin{equation}
	\left\{\begin{array}{l}\theta _{i} \left ( t+1 \right ) \sim N\left (\mu ,\sigma^{2} \right ) ,\theta =\theta _{i}\left ( t \right ) \vspace{1ex} 
		\\\mu = \frac{ {\textstyle \sum_{i=1}^{N}} \theta _{\text {best} } }{N}\vspace{1ex}
		\\ {\vspace{1ex}} \sigma ^{2} = \frac{ {\textstyle \sum_{i=1}^{N}}\left ( \theta_{\text{best}}- \mu   \right )^{2} }{N}\vspace{1ex} 
	\end{array}\right.
\end{equation}
where $\mu$ is the expectation and $\sigma ^{2}$ is the variance.

\subsection{Algorithmic Framework}
The algorithm's pseudocode and flow will be explained below to understand ISL better.

\subsubsection{Pseudo Code}
The ISL algorithm's pseudocode is broken into three parts to make it more modular and easier to understand. Algorithm 1 (Intelligent Social Learning) is the main framework that initializes the agents' population, evaluates fitness, and implements the three strategy styles. Algorithm 2 (Function Evaluation) is a sub-function used to evaluate the fitness value of each agent in the population. Algorithm 3 (Style Phase) is also a sub-function that implements the three strategy styles: learning, imitation, and self-study. The algorithm is simpler to modify or enhance particular aspects without affecting its overall structure by segmenting it into these three parts.

In Algorithm 1, the input is learning style $\boldsymbol{\mu}_{\mathrm{pop} \_\text{A}}^{\theta}$ and its size $popA\_num$, imitation style $\boldsymbol{\mu}_{\mathrm{pop} \_\text{B}}^{\theta}$ and its size $popB\_num$, and self-study style $\boldsymbol{\mu}_{\mathrm{pop} \_\text{C}}^{\theta}$ and its size $popC\_num$. The sum of the three styles is the population $\boldsymbol{\mu}_{\mathrm{pop}}^{\theta}$ and its size $pop\_num$. The output is the best fitness value and corresponding $\mu _{\mathrm{best} }^{\theta }$. The neural network structure, the benchmark test cases, the total iteration steps, and the initial sampling must be set up before the algorithm can be run. In this paper, weight and bias terms are the parameters that need to be updated using a fully connected deep neural network during the iterative process. The number of iteration steps can be set differently according to benchmark cases because some benchmark cases have many states, while others have few states. The number of states increases the dimensionality, making it harder for the algorithm to search. The initial sampling uses a standard normal distribution to make the population more diverse. During each generation of training, the best result is put through multiple tests, and the average score is recorded. This protocol is implemented to improve the algorithm's robustness and shield the reported metrics from any bias of the population size. $\boldsymbol{R}$ is a storage space used to store the best agent in training generation to facilitate the subsequent selection of historical samples.

\renewcommand{\algorithmicrequire}{ \textbf{Input:}} 
\renewcommand{\algorithmicensure}{ \textbf{Output:}}
\begin{algorithm}
	\caption{Intelligent Social Learning}
	\label{alg:alg1}
	\begin{algorithmic}[1]
		\REQUIRE 
		Learning style	$\boldsymbol{\mu}_{\mathrm{pop} \_\text{A}}^{\theta}$, $\boldsymbol{\mu}_{\mathrm{pop} \_\text{A}}^{\theta}$'s size $popA\_num$;\\
		\hspace{0.80cm}Imitation style $\boldsymbol{\mu}_{\mathrm{pop} \_\text{B}}^{\theta}$, $\boldsymbol{\mu}_{\mathrm{pop} \_\text{B}}^{\theta}$'s size $popB\_num$;\\
		\hspace{0.80cm}Self-study style $\boldsymbol{\mu}_{\mathrm{pop} \_\text{C}}^{\theta}$, $\boldsymbol{\mu}_{\mathrm{pop} \_\text{C}}^{\theta}$'s size $popC\_num$;\\
		\hspace{0.80cm}The population $\boldsymbol{\mu}_{\mathrm{pop}}^{\theta}= \left \{ \boldsymbol{\mu}_{\mathrm{pop} \_\text{A}}^{\theta},\boldsymbol{\mu}_{\mathrm{pop} \_\text{B}}^{\theta},\boldsymbol{\mu}_{\mathrm{pop} \_\text{C}}^{\theta}\right \}$, $\boldsymbol{\mu}_{\mathrm{pop}}^{\theta}$'s size $pop\_num$.
		\ENSURE
		{\vspace{0.1cm}} The best fitness and corresponding $\mu _{\mathrm{best}}^{\theta }$
		\IF {$step< max\_step$}
		\IF {$i< sampling\_num$}
		\FOR {$j=1$ \TO $pop\_num$}
		\STATE Normal distribution sampling $\mu_{\mathrm{pop}}^{\theta} $, $\theta \sim N\left ( 0,1 \right ) $
		\STATE $fitness,t= {\mathrm{Evaluate}}\left (\mu_{\mathrm{pop}}^{\theta }\right)$, $step\mathrel{+}= t$
		\ENDFOR
		\STATE Sort $fitness$ and choose the current best $\mu _{\mathrm{best}}^{\theta}$ from $\mu_{\mathrm{pop}}^{\theta}$
		\FOR {$k=1$ \TO $test\_num$}
		\STATE $fitness,\_= {\mathrm{Evaluate}}\left (\mu_{\mathrm{best}}^{\theta }\right)$, ${test\_fitness}\mathrel{+}= fitness$
		\ENDFOR 
		\STATE ${test\_fitness}= {test\_fitness}/{test\_num}$, $i=i+1$
		\STATE Append $\mu _{\mathrm{best}}^{\theta}$ and $test\_fitness$ to the storage space $\boldsymbol{R}$, $\left ( \mu _{\mathrm{best}}^{\theta}, test\_fitness \right ) \in \boldsymbol{R}$
		\ELSE
		\FOR {$n=1$ \TO $popA\_num$}
		\STATE Update $\theta$ by learning style, $\mu_{\mathrm{pop} \_\text{A}}^{\theta }= {\mathrm{Learn}}\left (\mu ^{\theta } \right)$
		\STATE $fitness,t= {\mathrm{Evaluate}}\left (\mu_{\mathrm{pop} \_\text{A}}^{\theta }\right)$, $step\mathrel{+}= t$
		\ENDFOR 
		\FOR {$p=1$ \TO $popB\_num$}
		\STATE Update $\theta$ by imitation style,  $\mu_{\mathrm{pop} \_\text{B}}^{\theta }= {\mathrm{Imitate}}\left (\mu ^{\theta } \right)$ 
		\STATE $fitness,t= {\mathrm{Evaluate}}\left (\mu_{\mathrm{pop} \_\text{B}}^{\theta }\right)$, $step\mathrel{+}= t$
		\ENDFOR 
		\FOR {$q=1$ \TO $popC\_num$}
		\STATE Update $\theta$ by self-study style, $\mu_{\mathrm{pop} \_\text{C}}^{\theta }= {\mathrm{Self \mbox{-} study}}\left (\mu ^{\theta } \right)$ 
		\STATE $fitness,t= {\mathrm{Evaluate}}\left (\mu_{\mathrm{pop} \_\text{C}}^{\theta }\right)$, $step\mathrel{+}= t$ 
		\ENDFOR 
		\STATE Sort $fitness$ and choose the current best $\mu _{\mathrm{best}}^{\theta} $ from $\left \{ \boldsymbol{\mu}_{\mathrm{pop} \_\text{A}}^{\theta},\boldsymbol{\mu}_{\mathrm{pop} \_\text{B}}^{\theta},\boldsymbol{\mu}_{\mathrm{pop} \_\text{C}}^{\theta}\right \}$ 
		
		\FOR {$k=1$ \TO $test\_num$}
		\STATE $fitness,\_= {\mathrm{Evaluate}}\left (\mu_{\mathrm{best}}^{\theta }\right)$, ${test\_fitness}\mathrel{+}= fitness$
		\ENDFOR 
		\STATE ${test\_fitness}= {test\_fitness}/{test\_num}$, $step\mathrel{+}= step$
		\STATE Append $\mu _{\mathrm{best}}^{\theta}$ and $test\_fitness$ to the storage space $\boldsymbol{R}$, $\left ( \mu _{\mathrm{best}}^{\theta}, test\_fitness \right ) \in \boldsymbol{R}$ 
		\ENDIF
		\ENDIF
	\end{algorithmic}
\end{algorithm}

In Algorithm 2, the stochastic policy is used for continuous control. The population of agents is then evaluated in an interaction episode with the environment. The fitness for each agent is computed as the cumulative sum of the rewards they receive over the timesteps in that episode. Using a fitness metric that consolidates returns across an entire episode makes ISL indifferent to the sparsity of reward distribution and robust to long time horizons. To make the comparisons fair across single agents and population-based ISL algorithms, the total steps taken by all agents in the population are cumulative. For instance, an episode of a benchmark case consists of 1000 steps, and a population of 10 agents would incur 10,000 steps per iteration. The algorithm reports the average with error bars logging the standard deviation after running five independent statistical runs with varying random seeds.

\renewcommand{\algorithmicrequire}{\textbf{procedure}}
\renewcommand{\algorithmicensure}{\textbf{end procedure}}
\begin{algorithm}[h]
	\caption{Function Evaluation}
	\label{alg:alg2}
	\begin{algorithmic}[1]
		\REQUIRE $\mathrm{Evaluate}\left ( \mu ^{\theta }  \right )$
		\STATE $fitness = 0$, $t=0$
		\STATE  Reset the environment and get the initial state $s$
		\WHILE  {env is not done} 
		\STATE	Stochastic policy for continuous control, compute mean and log variance using the neural network, $\hat{\mu}  = \boldsymbol{\mu}\left ( s; \boldsymbol{\theta}^{\upmu} \right )$ and $\hat{\rho }  = \boldsymbol{\rho }\left (s; \boldsymbol{\theta}^{\uprho} \right)$
		\STATE Compute $\hat{\sigma } ^{2} = \exp \left ( \hat{\rho}\right)$
		\STATE Randomly sample action $a_{t}$ by $a\sim N\left ( \hat{\mu} ,\hat{\sigma } ^{2}  \right)$ 
		\STATE Execute action $a_{t}$ and observe reward $r_{t}$ and new state $s_{t+1}$, $r_{t} = {\mu _{pop}}^{\theta} \left ( a_{t}\mid s_{t} \right) $ 
		\STATE $fitness=fitness+ r_{t}$, $s=s_{t+1}$, and $t=t+1$ 
		\ENDWHILE 
		\STATE \textbf{return} $fitness$, $t$ 
		\ENSURE
	\end{algorithmic}
\end{algorithm}

For updating network parameters, three methods are utilized in Algorithm 3. The lower and upper bounds for the best historical agent are determined by selecting the storage space $\boldsymbol{R}$ agent with the highest fitness value. Using Eqs. (8) and (9), the Lévy flight strategy, the learning style updates all agent paetersram. Imitation style updates all or some agent parameters using disturbance strategy by Eq. (11). Self-study style updates all agent parameters using normal distribution sampling by Eq. (12). Updated parameters are limited to 1.5 times the best agent's corresponding parameters to prevent excessive parameter updates.

\renewcommand{\algorithmicrequire}{\textbf{procedure}}
\renewcommand{\algorithmicensure}{\textbf{end procedure}}
\begin{algorithm}[h]
	\caption{Style Phase}
	\label{alg:alg3}
	\begin{algorithmic}[1]
		\REQUIRE $\mathrm{Learn}\left(\mu^{\theta }\right)$, $\mathrm{Imitate}\left(\mu ^{\theta }\right)$, and $\mathrm{Self\mbox{-} study}\left(\mu^{\theta}\right)$
		\STATE Sort $test\_fitness$ and choose the current best $\mu _{\mathrm{best}}^{\theta }$ from $\boldsymbol{R}$
		\STATE Get the lower and upper bounds of the current best $\mu _{\mathrm{best}}^{\theta }$, $\mu _{\mathrm{best}}^{\theta}= \left [ \boldsymbol{L}_{b}, \boldsymbol{U}_{b} \right ]$ 
		\STATE $\mu_{\mathrm{pop} \_\text{A}}^{\theta }= \mathrm{Learn}\left(\mu^{\theta }\right)$ $\leftarrow$ Update all $\theta $ of $\mu ^{\theta}$ using Lévy flight strategy by Eqs. (8) and (9) in learning style
		\IF {Pr $\ge$ rand(0, 1)}
		\STATE $\mu_{\mathrm{pop} \_\text{B}}^{\theta }=\mathrm{Imitate}\left(\mu ^{\theta }\right)$ $\leftarrow$ Update all $\theta$ of $ \mu ^{\theta}$ using disturbance strategy by Eq. (11) in imitation style
		\ELSE
		\STATE $\mu_{\mathrm{pop} \_\text{B}}^{\theta }=\mathrm{Imitate}\left(\mu ^{\theta }\right)$ $\leftarrow$ Update some $\theta$ of $\mu ^{\theta}$ using disturbance strategy by Eq. (11) in imitation style
		\ENDIF
		\STATE $\mu_{\mathrm{pop} \_\text{C}}^{\theta }=\mathrm{Self\mbox{-} study}\left(\mu^{\theta}\right)$ $\leftarrow$ Update all $\theta$ of $\mu ^{\theta}$ using normal distribution sampling by Eq. (12) in self-study style
		\STATE Limit the range of $\theta$, clamp\\$\left \{ \left ( {\mu}_{\mathrm{pop} \_\text{A}}^{\theta},{\mu}_{\mathrm{pop} \_\text{B}}^{\theta},{\mu}_{\mathrm{pop} \_\text{C}}^{\theta} \right ) ,\text{min}= 1.5\boldsymbol{L}_{b}, \text{max}= 1.5\boldsymbol{U}_{b}\right \}$ 
		\STATE  \textbf{return} ${\mu}_{\mathrm{pop} \_\text{A}}^{\theta}$, ${\mu}_{\mathrm{pop} \_\text{B}}^{\theta}$, ${\mu}_{\mathrm{pop} \_\text{C}}^{\theta}$ 
		\ENSURE
	\end{algorithmic}
\end{algorithm}

\subsubsection{Algorithmic Flow}
To demonstrate the whole optimization flow, an illustration with specific algorithm steps is given in Fig. 4. Set the sizes of the learning, imitation, and self-study styles after creating a neural network and initializing the associated parameters. The initial population's network parameters adopt standard normal distribution to satisfy the total number of iterations and the number of initial sampling. Initialize the task environment, obtain the initial state $s_{0}$, and use stochastic policy for continuous control to obtain the action $a_{t}$ in each state $s_{t}$. The agent interacts with the environment to get the reward value $r_{t}$ in the current state and the new state $s_{t+1}$ and gets the fitness value $f$ until the end of the episode. In each iteration, sort the agents' fitness values to obtain the network model parameters corresponding to the highest current fitness value score, verified by multiple average tests. That is the optimal value in the current iteration. Through multiple average tests, return the average fitness value $test\_fitness$ and corresponding network model parameters and initial sampling numbers $i$ and the total iteration numbers $step$. The algorithm adopts three strategic styles when the initial sampling numbers and the total iteration steps are met. Find the network model corresponding to the highest historical average test fitness value, the historical optimal value. In the learning style, update network parameters refer to Eqs. (8) and (9). In the imitation style, update network parameters refer to Eq. (11). In the self-study style, update network parameters refer to Eq. (12). Until the end of the episode, repeat the process of the aforementioned agents interacting with the environment to determine the fitness value $f$ of the episode and the action and reward values in each state. When the total number of iterations has been reached, the iterative process loops until the final controller for the task is the model corresponding to the obtained historical optimal value.

\begin{figure*}[h]
	\centering
	\includegraphics[width=0.8\textwidth]{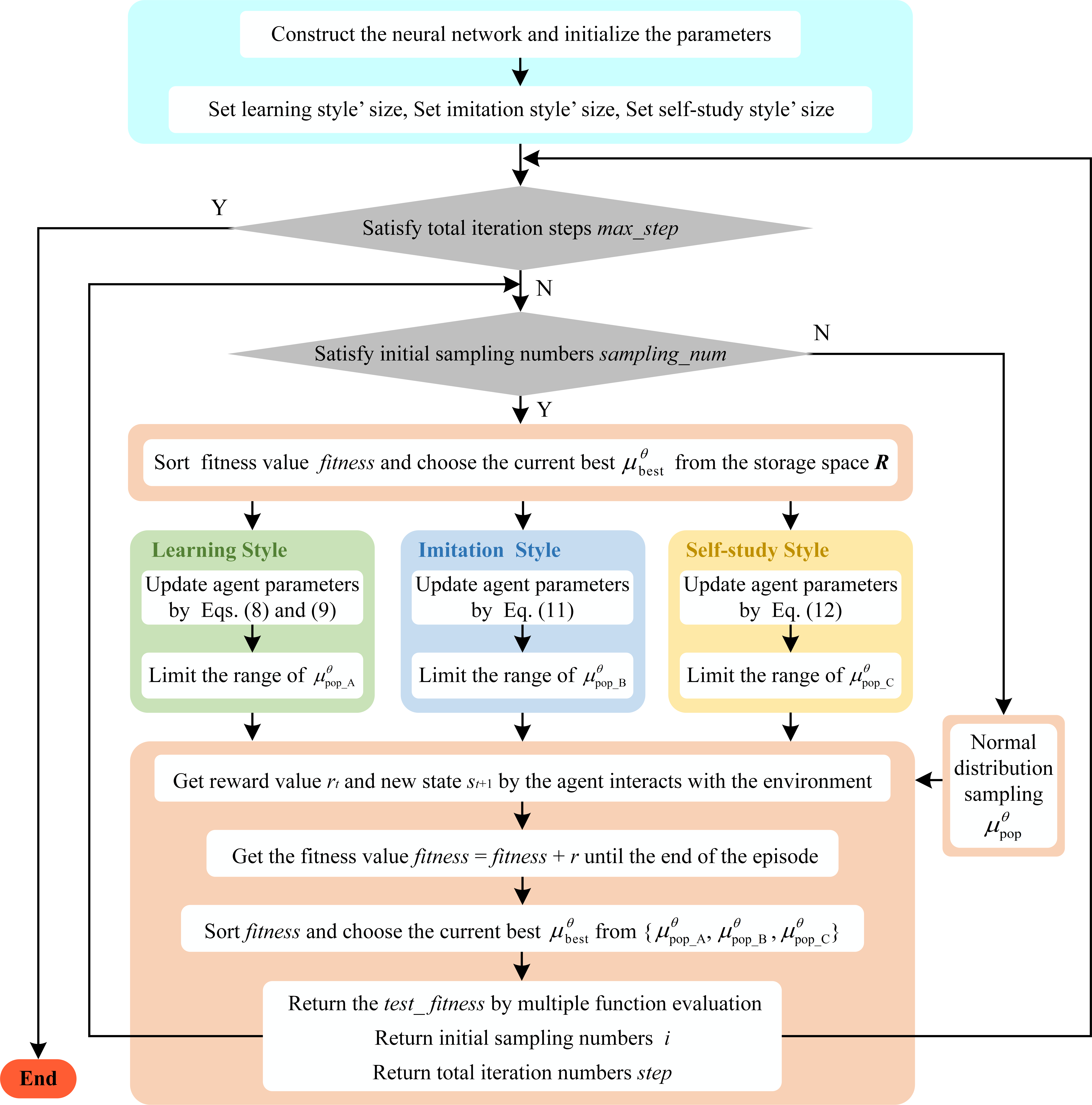}
	\caption{Overall optimization flow of ISL.}
	\label{fig_4}
\end{figure*}

\section{Evaluation and Application}
\subsection{Benchmark Cases}
To verify the efficiency and robustness of ISL, six continuous control benchmark cases simulated using MuJoCo are given for testing, and four methods are used for comparison. MuJoCo is a physics engine that makes it easier to research and develop in fields like robotics, biomechanics, graphics, animation, and other fields where quick and accurate simulations are needed\cite{todorov2012mujoco}. Fig. 5 shows the six benchmark cases in MuJoCo: Half Cheetah, Swimmer, Hopper, Ant, Walker2D, and Reacher\cite{duan2016benchmarking}. These benchmark cases have different observation shapes and action spaces and are widely used for algorithm tests.

\begin{figure}[h]
	\centering
	\includegraphics[width=3.0 in]{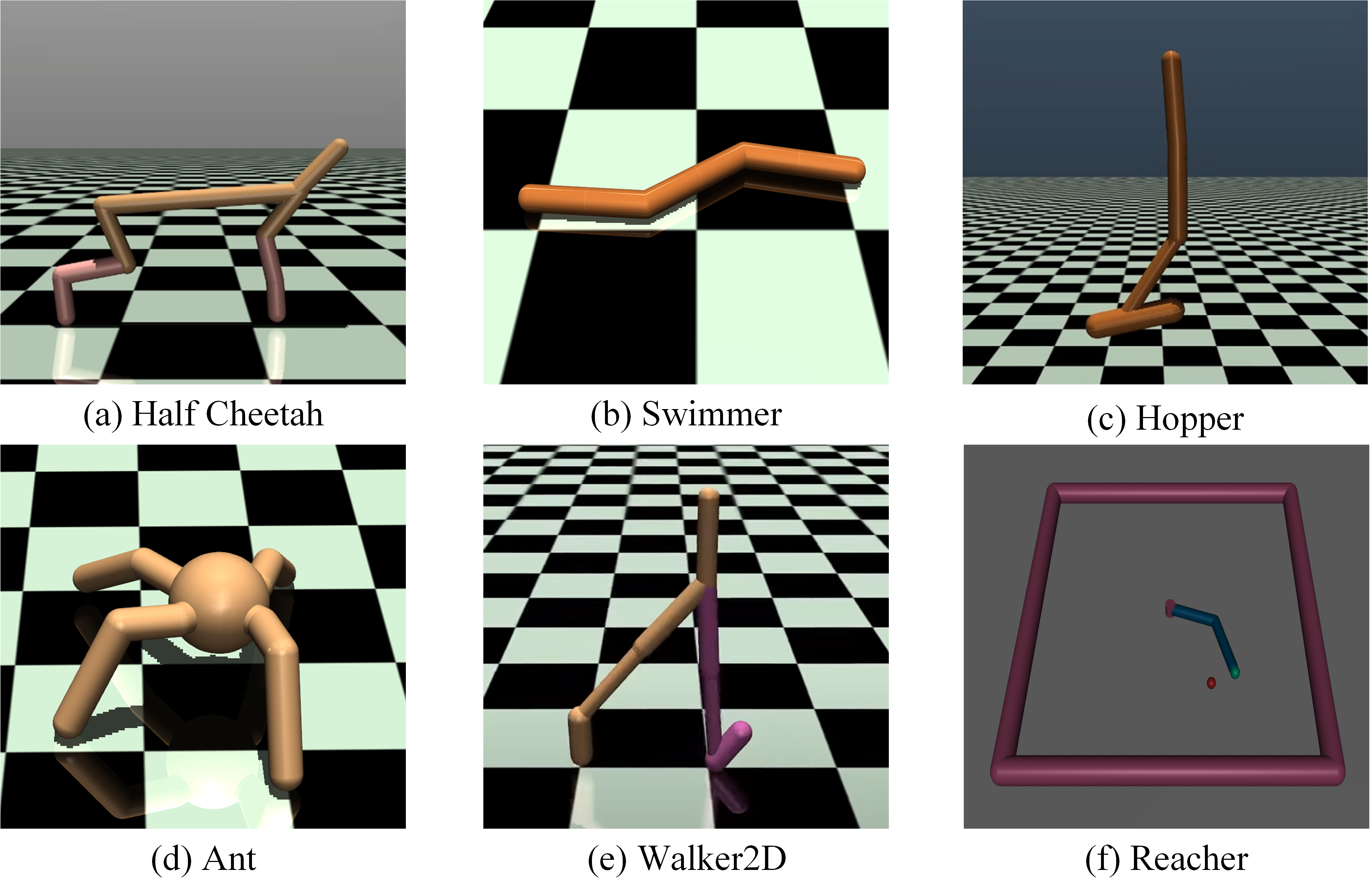}
	\caption{The benchmark cases in MuJoCo.}
	\label{fig_5}
\end{figure}

The four algorithms compared with ISL are DDPG, SAC, PPO, and EA. DDPG, SAC, and PPO belong to the reinforcement learning domain, with DDPG and SAC being typical off-policy deep reinforcement learning algorithms, while PPO is on-policy. SAC's stochastic strategy is the primary distinction between DDPG and SAC. DDPG employs a deterministic strategy. EA is in the field of Neuroevolution, which is a class of black-box optimization techniques inspired by natural evolution.

\subsection{Calculation Results}
ISL requires a few hyperparameters to be adjusted. In this simulation, the number of populations was set to 10, among which the number of populations $n$ in the learning style was set to 5, the number of populations $p$ in the imitation style was set to 3, and the number of populations $q$ in self-study style was set to 2. Based on most Lévy flight experience, the proportional factor of the step size $\alpha_{\max}$ was set to 0.1, and the $\alpha_{\min} $ was set to 0.01. The perturbation magnitude $a$ in imitation style was set to -1, and $b$ was set to 1. OpenAI Baselines\cite{dhariwal2017openai} implemented DDPG, SAC, and PPO, and their hyperparameters were slightly modified for enhanced performance. A standard Genetic Algorithm\cite{khadka2018evolution} was used to implement EA, and its hyperparameters were set to match the original papers. A computer with the Core i7-10700 CPU (2.90GHz) was used for all tests.

With solid curves representing mean values and shaded regions indicating standard deviations, Fig. 6 depicts the performance results of each algorithm on various benchmark cases. The results show that different algorithms perform better or worse on different benchmark cases, indicating that each algorithm has strengths and weaknesses depending on the task. DDPG performs best on Half Cheetah but is not outstanding on other benchmark cases. EA has no advantage in other benchmark cases except for Swimmer compared to ISL. PPO performs fairly well in Hopper and Reacher, slightly better than ISL in Hopper, but mediocre in the other benchmark cases. ISL performs best on Swimmer and Reacher and well on other benchmark cases. ISL's fitness value grows stepwise during the training process and can quickly converge. This is because all agents learn from the best agent as a reference, resulting in a more stable learning process with lower variance.

\begin{figure*}[htb]
	\centering
	\small
	\includegraphics[width=0.8\textwidth]{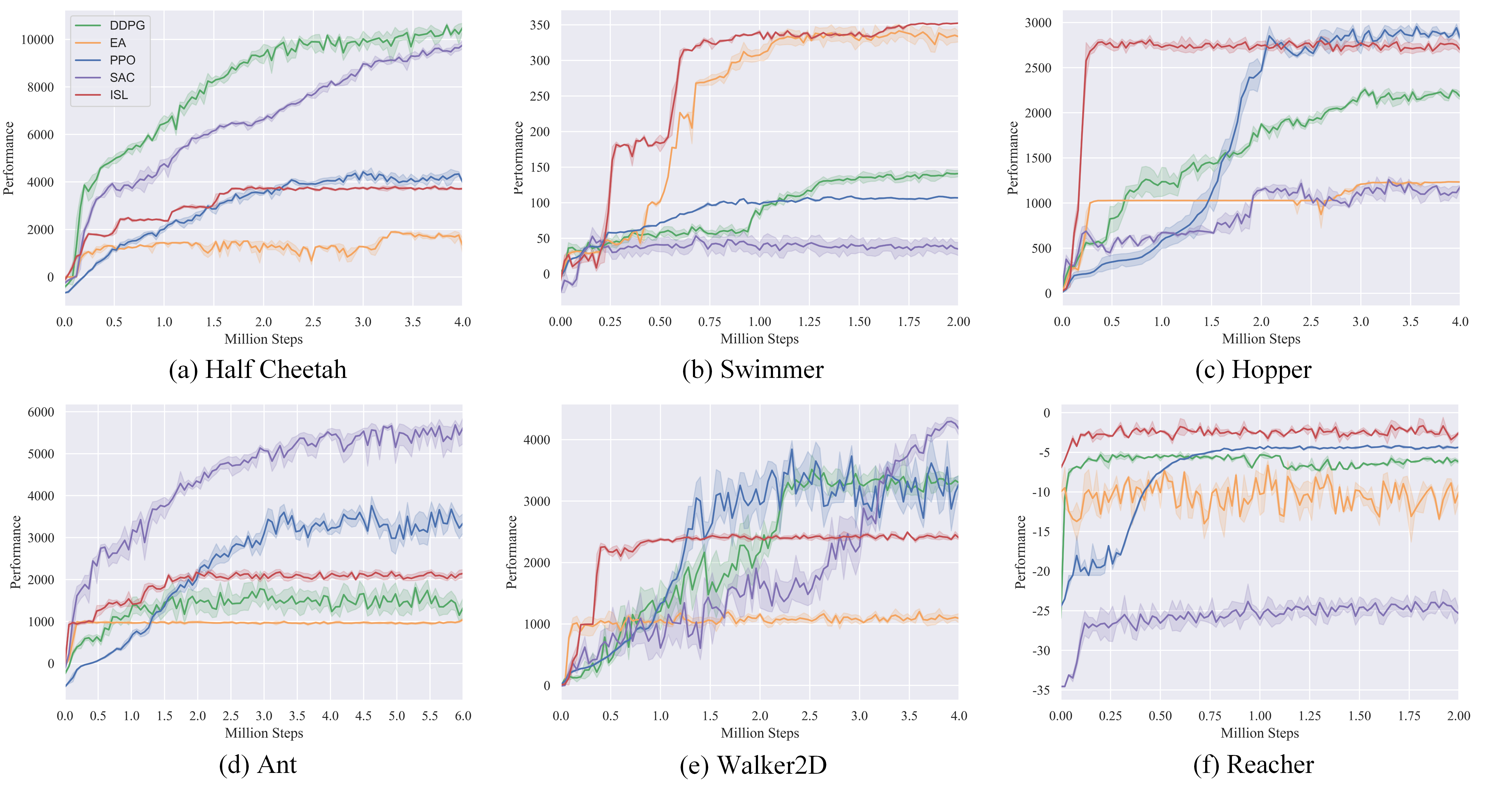}
	\caption{Learning curves for continuous control benchmark cases on MuJoCo.}
	\label{fig_6}
\end{figure*}

\begin{table*}[h]
	\caption{Maximum performance and computing time.\label{tab:table1}}
	\centering
	\resizebox{\linewidth }{!}{
		\begin{tabular}{cccccccc}
			\specialrule{1pt}{0pt}{0pt} 
			\multirow{3}{*}{ \textbf{Algorithms}} & \textbf{Benchmarks}   & \textbf{HalfCheetah} & \textbf{Swimmer}    & \textbf{Hopper}      & \textbf{Ant}         & \textbf{Walker2d}    & \textbf{Reacher}    \\ \cline{2-8}\noalign{\smallskip}
			& State/Action & 17/6        & 8/2        & 11/3        & 27/8        & 17/6        & 11/2       \\ \cline{2-8}\noalign{\smallskip}
			& Step/ mil.   & 4           & 2          & 4           & 6           & 4           & 2          \\ \specialrule{1pt}{0pt}{0pt}
			\multirow{2}{*}{\textbf{DDPG}}       & Maximum      & \textbf{10800.67(1)} & 146.74(3)  & 2296.12(3)  & 2229.42(4)  & 3796.61(3)  & -4.87(3)   \\
			& Time/min     & 1075{[}4{]} & 506{[}4{]} & 1105{[}4{]} & 2125{[}4{]} & 1165{[}4{]} & 495{[}4{]} \\
			\multirow{2}{*}{\textbf{SAC}}        & Maximum      & 10012.22(2) & 65.47(5)   & 1304.93(4)  & \textbf{5988.86(1)}  & \textbf{4402.99(1)}  & -21.62(5)   \\
			& Time/min     & 1563{[}5{]} & 684{[}5{]} & 1509{[}5{]} & 2686{[}5{]} & 1521{[}5{]} & 786{[}5{]} \\
			\multirow{2}{*}{\textbf{PPO}}        & Maximum      & 4636.03(3)  & 109.63(4)  & \textbf{3012.16(1)}  & 4165.46(2)  & 4160.82(2)  & -4.05(2)   \\
			& Time/min     & 165{[}3{]}  & 54{[}3{]}  & 145{[}3{]}  & 475{[}3{]}  & 155{[}3{]}  & \textbf{53{[}1{]}}  \\
			\multirow{2}{*}{\textbf{EA}}         & Maximum      & 1948.76(5)  & 350.85(2)  & 1235.58(5)  & 1097(5)     & 1289.27(5)  & -6.02(4)   \\
			& Time/min     & 25{[}2{]}   & 20{[}2{]}  & 69{[}2{]}   & 153{[}2{]}  & 87{[}2{]}   & 75{[}3{]}  \\
			\multirow{2}{*}{\textbf{ISL}}        & Maximum      & 3908.96(4)  & \textbf{354.49(1)}  & \textbf{2901.73(2)}  & 2365.39(3)  & 2534.69(4)  & \textbf{-1.57(1)}   \\
			& Time/min     & \textbf{20{[}1{]}}   & \textbf{16{[}1{]}}  & \textbf{57{[}1{]}}   & \textbf{131{[}1{]}}  & \textbf{74{[}1{]}}   & 67{[}2{]} \\
			\specialrule{1pt}{0pt}{0pt} 
	\end{tabular}}
\end{table*}

The maximum performance and computing time of the test results are shown in Table 1 for further quantitative analysis, where "( )" indicates the ranking of performance and "[ ]" indicates the ranking of computing time. It can be seen from the table that SAC won first place (Rank 1) on both Ant and Walker2d and won second place (Rank 2) on HalfCheetah. However, the performance difference between SAC and the first-place DDPG on HalfCheetah is not much. ISL won first place (Rank 1) in both Swimmer and Reacher, and its performance is far better than DDPG, SAC, and PPO. Similarly, the performance difference between ISL and the first-place PPO in Hopper is not too big. Compared to the HalfCheetah, Ant, and Walker2d cases, the primary distinction of the Swimmer and Reacher lies in their smaller state shape and action space and their structure belonging to the series form. Therefore, the performance of ISL is not outstanding in high-dimensional problems, and it shares similar drawbacks with intelligent optimization algorithms, such as a susceptibility to getting stuck in local optima.

It is worth noting that ISL can save computing resources and reduce computing time during the simulation. In five benchmark cases, SAC beat out other algorithms to win four first places (Rank 1) and one second place (Rank 2), all in significantly less time. For example, under the same hardware conditions, ISL reaches 4 million steps in HalfCheetah in about 20 minutes, while DDPG is about 54 times that of ISL, SAC is about 78 times that of ISL, and PPO is about eight times that of ISL. However, EA and ISL take roughly the same amount of time to solve because they can use multithreaded parallel computing and do not require gradient updates. In addition, compared with EA, ISL can use the three strategies of learning style, imitation style, and self-study style to balance the relationship between exploration and exploitation, effectively expand the global search range, and is slightly easier to jump out the local optima. If ISL takes the same time as other algorithms, it should show better results.

Based on the above discussion and analysis, it can be concluded that ISL demonstrates superior comprehensive results in terms of performance and time consumption compared to other algorithms. Furthermore, ISL exhibits its own unique characteristics and advantages, and has certain research value and promotion significance.

\subsection{Engineering Application}
To further verify the algorithm's effectiveness and practical engineering value, ISL is used for robot object grasping. The robot used in this study is the 6-degree-of-freedom UR3 robot. Despite its compact size, the UR3 boasts impressive functionality, with a payload capacity of 3kg and the ability to perform operations within a radius of 500mm. All wrist joints of this desktop collaborative robot can rotate ± 360 degrees, and the end joints can rotate infinitely, allowing it to perform high-precision and light assembly tasks easily. 

As shown in Fig. 7, the UR3 robot simulation environment is built in MuJoCo. In the figure, three cube blocks on the desktop are target objects, and the yellow target dot on the other side of the desktop is a random location. The task requirement is to grasp the three cube blocks and place them at the target dot. The input state dimension of the entire environment is 9, consisting of 3 distance information in space and 6 joint angle information of the robot. The output action space consists of 6 actions, representing the angle of each joint of the 6 DOF robot.

\begin{figure}[htb]
	\centering
	\includegraphics[width=3.0 in]{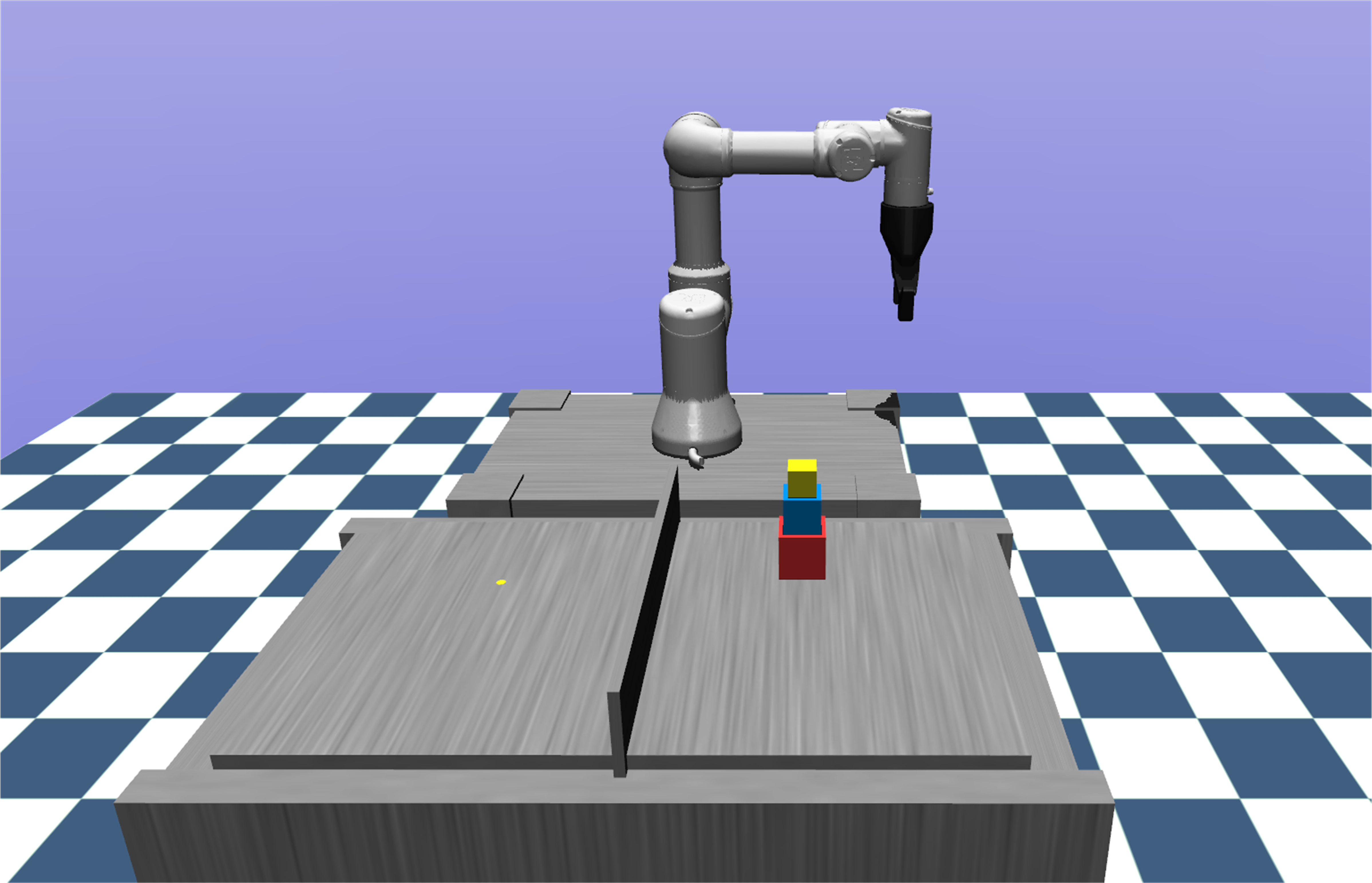}
	\caption{The simulation environment of UR3 in MuJoCo.}
	\label{fig_7}
\end{figure}

\begin{figure*}[htb]
	\centering
	\includegraphics[width=0.8\textwidth]{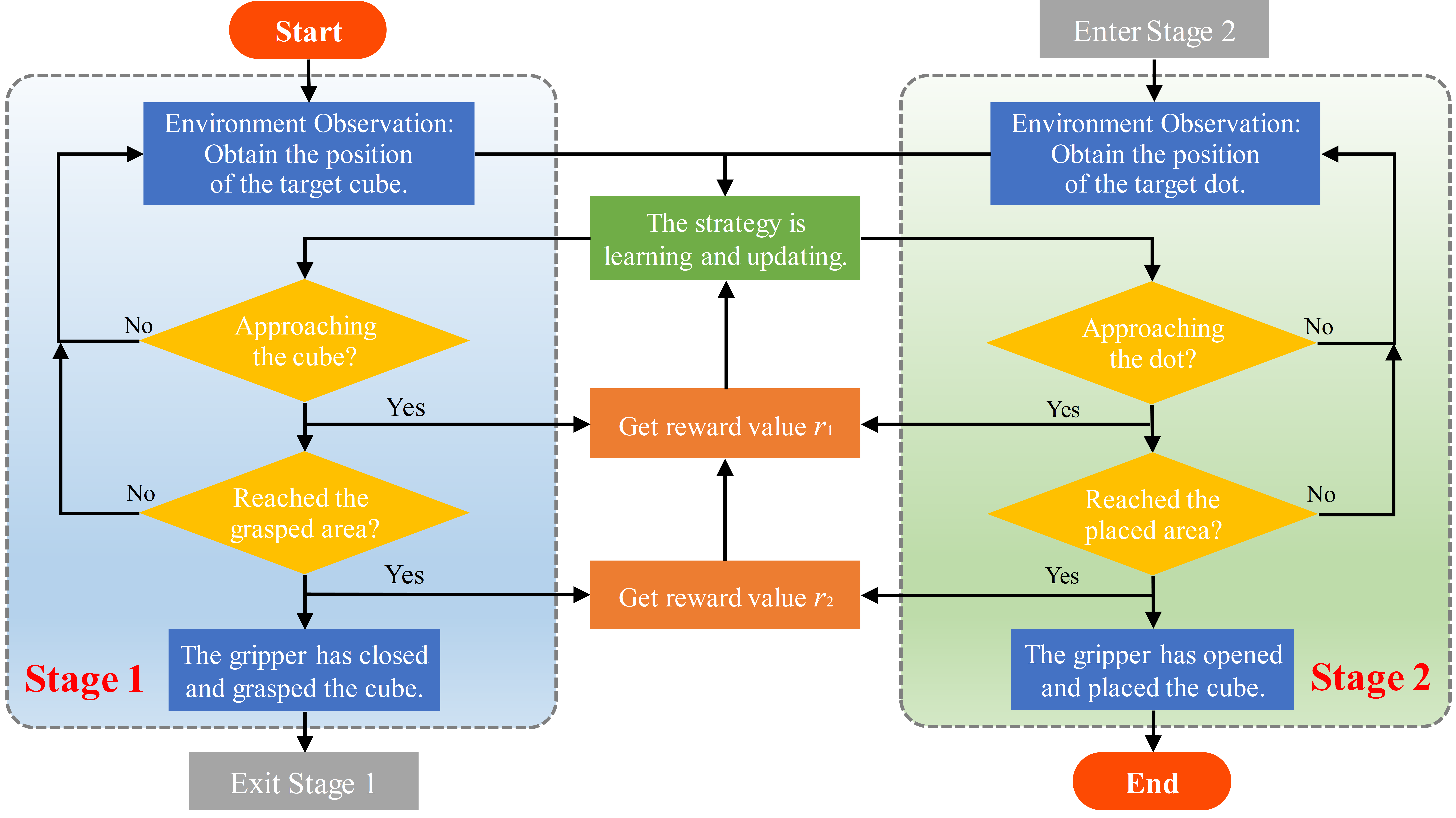}
	\caption{Robot grasping task flowchart.}
	\label{fig_8}
\end{figure*}
The task assigned to the robot consists of two stages: As depicted in Fig. 8, Stage 1 entails grasping the target object, and Stage 2 entails placing the target object. Stage 1 must be completed first throughout the task before Stage 2. Due to the common tasks in Stages 1 and 2, the two stages share control strategies, but the reward function is designed separately. The design of rewards is an important part of the learning process because it helps algorithms achieve their goals more quickly and reliably. Therefore, according to the robotic assignment task process, the reward function is set as a composite reward function $R$, consisting of a guided reward function $r_{1}$ and a sparse reward function $r_{2}$.

Approaching the target position: The robot gradually moves toward the target position from its initial location. A guided reward function $r_{1}$ is designed to guide the robot's end gripper to the target position, as shown in Eq. (13). The target object's location is the target position in Stage 1, while the target dot's location is the target position in Stage 2.

\begin{equation}
	{r_{1}} = \begin{cases}
		-\left \| P_{\text{robot}}- P_{\text{cube}} \right \| _{2},&{\text{State1}} \\ 
		-\left \| P_{\text{robot}}- P_{\text{dot}} \right \| _{2},&{\text{State2.}} 
	\end{cases}
\end{equation}
where $P_{\text{robot}}$ refers to the position of the robot's end effector, $P_{\text{cube}}$ refers to the position of the target object, and $P_{\text{dot}}$ refers to the position of the target dot. Stage 1 indicates the process of executing Stage 1 at this time, while Stage 2 indicates the process of executing Stage 2 at this time.

Performing task actions: The robot gradually moves from the target position to the executable area. A sparse reward function  $r_2$ is defined to determine whether the robot's end gripper has completed the assignment task, as shown in Eq. (14). The task action in Stage 1 is to grasp the target object, whereas the task action in Stage 2 is to place the target object.

\begin{equation}
	{r_{2}} = \begin{cases}
		1, &\begin{cases}
			{\left| P_{\text{robot}}- P_{\text{cube}} \right |\le \bigtriangleup ^{\text{near1}}},&{\text{State1}} \\ 
			{\left | P_{\text{robot}}-  P_{\text{dot}} \right |\le  \bigtriangleup ^{\text{near2}}},&{\text{State2.}} 
		\end{cases}\\
		0,&{\text{otherwise.}}
	\end{cases}
\end{equation}
where $\bigtriangleup ^{\text{near1}}$ refers to the executable area of Stage 1, where the robot's end gripper executes a closing action. In contrast, $\bigtriangleup ^{\text{near2}}$ refers to the executable area of Stage 2, where the robot's end gripper executes an opening action. It is worth noting that the reachable area must simultaneously satisfy the conditions in all three directions of $x$, $y$, and $z$.

Eq. (15) demonstrates that the sum of the guided reward function $r_1$ and the sparse reward function $r_2$ is the composite reward function $R$.

\begin{equation}
	R = {r_1} + {r_2}
\end{equation}

The ISL algorithm's performance in object grasping by the UR3 robot is depicted in Fig. 9. The ISL algorithm exhibits strong exploration in the early training stages, leading to random test results. After numerous training sessions, the UR3 robot's end effector can accurately reach the intended location. After a long training period, the UR3 can accurately grasp the target object and quickly place it at the target dot, achieving the desired effect. As depicted in Fig. 9, there are three stages to the grasping process: in the first stage, target object 1 is grasped and placed successfully on the yellow target dot; in the second stage, target object 2 is grasped and placed successfully on the target object 1; in the third stage, the target object 3 is grasped and placed successfully on the target object 2.

\begin{figure*}[htb]
	\centering
	\includegraphics[width=0.8\textwidth]{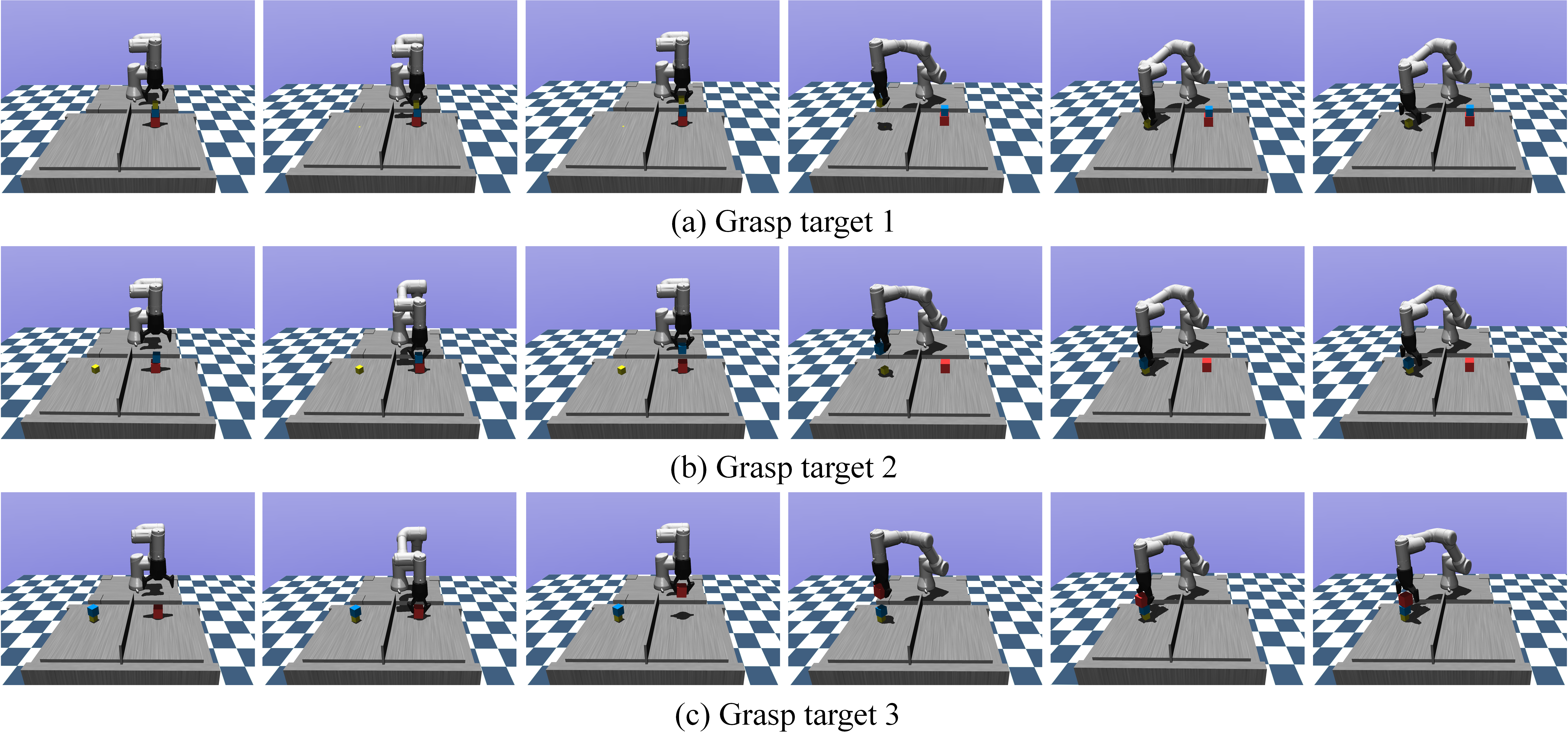}
	\caption{The simulation result of object grasping by the UR3.}
	\label{fig_9}
\end{figure*}

In this study, the excellent ISL model developed through simulation training is applied to the actual UR3 grasp control, and the experimental setup is depicted in Fig. 10. A UR3 robot, Robotiq 2-finger grippers, an Intel Realsense D435 camera, target wooden blocks, and a high-performance computer are the necessary components for the experiment. The target block is affixed with an AprilTag visual two-dimensional barcode label, facilitating the target object's identification and pose estimation.

\begin{figure}[htb]
	\centering
	\includegraphics[width=3.0 in]{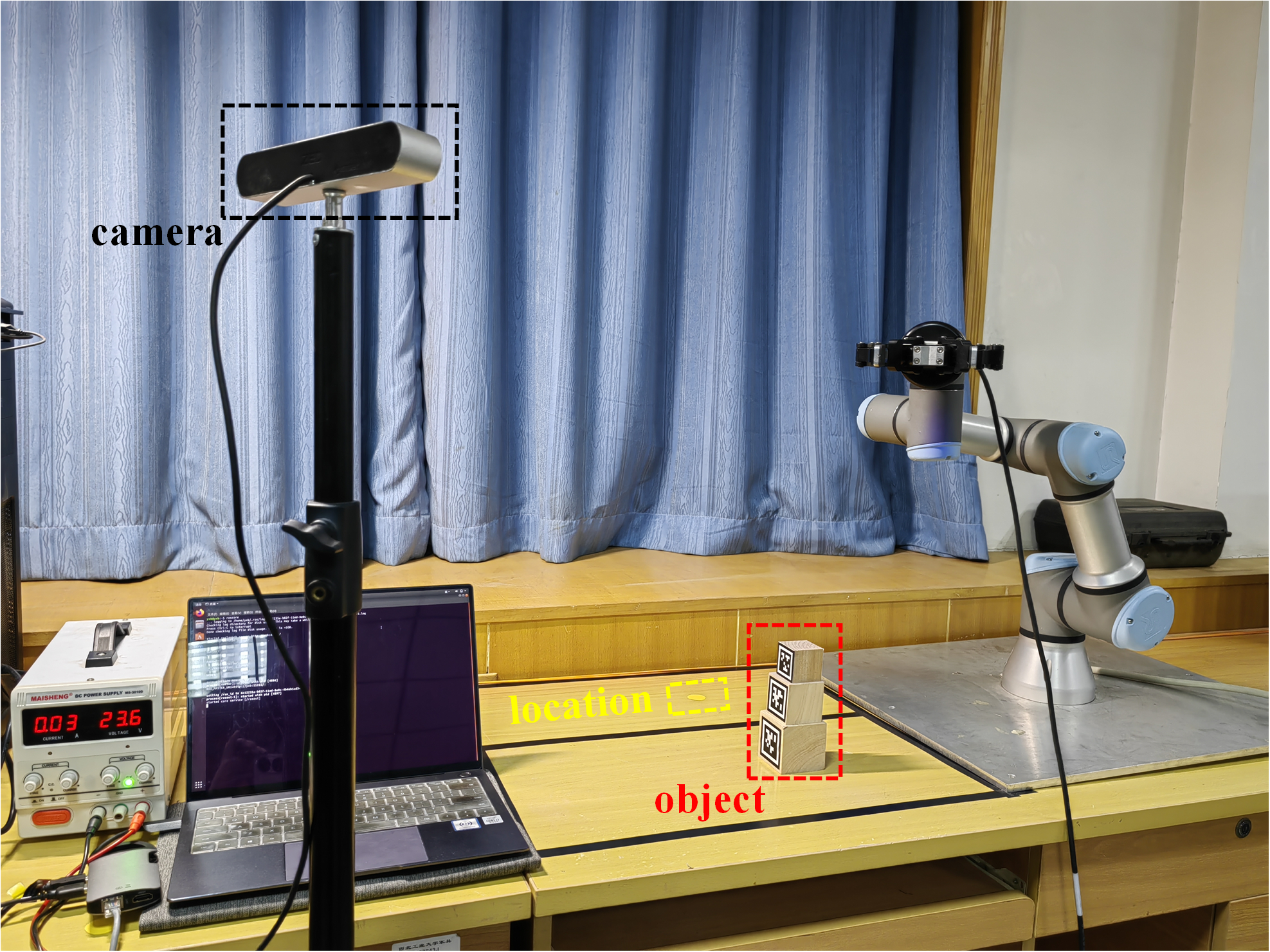}
	\caption{The experiment environment of UR3.}
	\label{fig_10}
\end{figure}

\begin{figure*}[htb]
	\centering
	\includegraphics[width=0.8\textwidth]{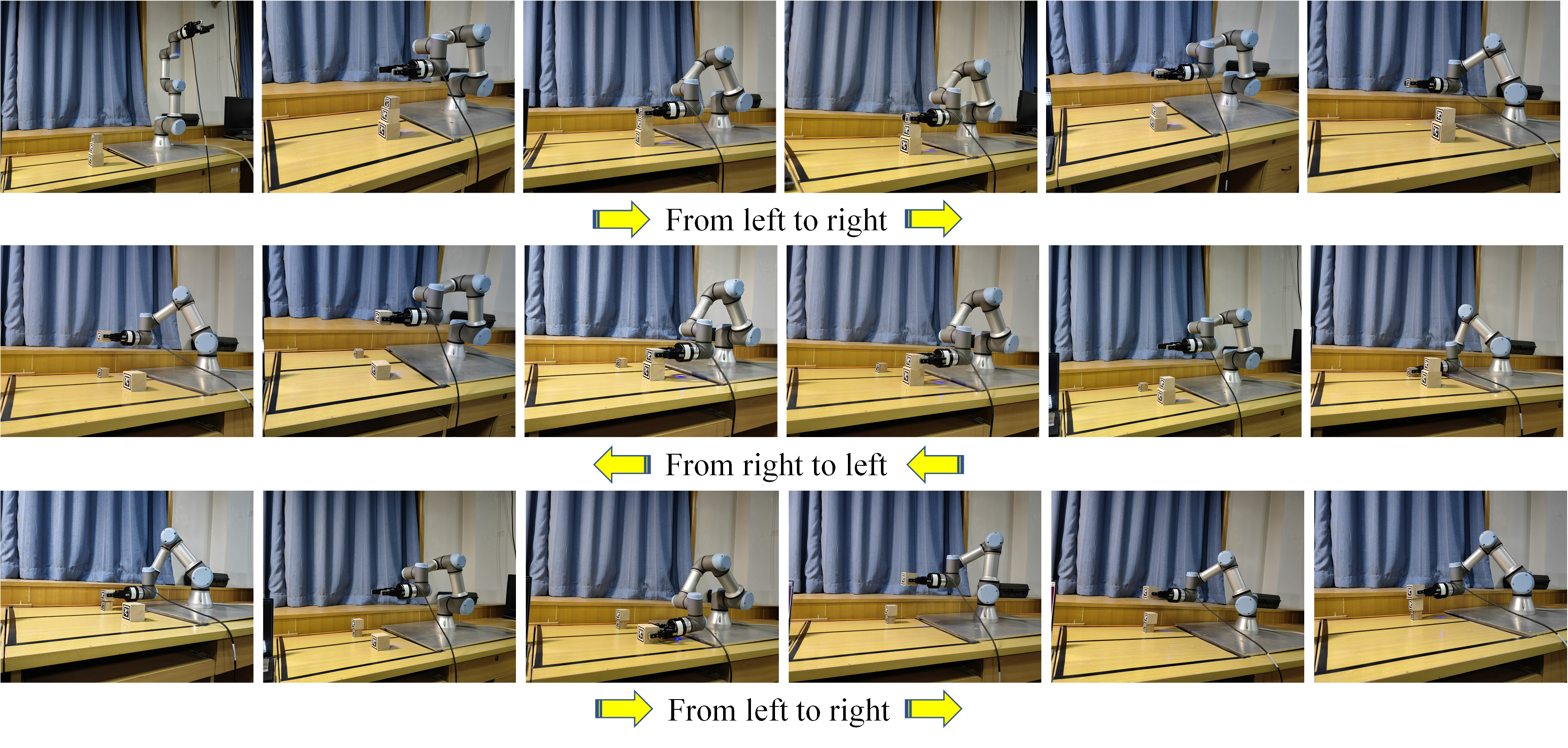}
	\caption{The experimental result of object grasping by the UR3.}
	\label{fig_11}
\end{figure*}

The depth camera is used in advance to obtain the target object and target dot's positional information. The joint angle values in the simulation environment are transmitted to the joint control of the actual robot via the network. The robot initially struggles to perform at its best because the real and simulated environments differ. However, after multiple training rounds, its performance gradually improves, and the experimental results are as anticipated, as depicted in Fig. 11.

\section{Conclusion}
This study proposes an efficient ISL algorithm to solve black-box robot control problems. ISL emulates mutual learning among individuals in human social groups, including learning style, imitation style, and self-study style. In summary, the significant characteristics of ISL are as follows: (1) ISL possesses powerful search capability. Its three strategic styles skillfully balance the relationship between exploration and exploitation while promoting population diversity, expanding the scope of global search, and facilitating escape from local optima. (2) ISL exhibits a fast iterative-solving ability. Multithreaded parallel computing can be employed during the simulation process without requiring gradient updates, conserving computing resources and reducing computational time. (3) ISL requires a few adjustable hyperparameters. A few hyperparameters need to be adjusted during the update iteration process, which is easy to converge quickly. (4) ISL boasts a high utilization rate of reward returns. It does not care about the issue of sparse rewards or environmental noise because it looks at the total returns for the entire episode.

To verify ISL's efficiency and robustness, six continuous control benchmark cases simulated using MuJoCo are used for testing and compared with the other four methods. The results show that ISL performs well in Swimmer and Reacher and is slightly inferior in other benchmark cases but not the worst, with significantly less computation time than other algorithms. Finally, ISL is applied to practical engineering, i.e., the UR3 robot is used to grasp and quickly place the target object accurately. Both simulation and experimental results have achieved satisfactory outcomes. Therefore, ISL has its characteristics and advantages, research value, and promotion significance.

Currently, ISL is capable of achieving intelligent motion control for most robots. However, when faced with robots that have complex multi-joint parallel structures, that is, high-dimensional problems, ISL reveals its limitations. This challenge is also encountered in the field of DRL and IO when dealing with high-dimensional optimization problems. In future work, there is an interest in expanding the capabilities of ISL to address the optimal control problem in complex robots.

\printcredits

\bibliographystyle{elsarticle-num}
\bibliography{ref}

\end{document}